\documentclass[runningheads]{llncs}

\usepackage{eccv}

\usepackage{eccvabbrv}
\usepackage{graphicx}
\usepackage{booktabs}

\usepackage[accsupp]{axessibility}
\usepackage{hyperref}

\usepackage{orcidlink}
\usepackage{booktabs}        
\usepackage{xcolor} 
\usepackage{multicol} 
\usepackage{multirow} 
\usepackage{amsmath} 
\usepackage{graphicx} 
\usepackage{bbding}   
\usepackage{makecell}   
\usepackage{diagbox} 
\usepackage{colortbl} 
\usepackage{amssymb}
\usepackage{algorithm} 
\usepackage{algorithmic}

\begin{document}

\title{ANFI: Rethinking Neighbor Feature Interaction in Person Re-ID}

\author{Xulin Li\inst{1,2} \and
Yan Lu\inst{3} \and
Bin Liu\inst{1,2}\thanks{Corresponding author.} \and
Jiaze Li\inst{1,2} \and
Qinhong Yang\inst{1,2} \and\\
Tao Gong\inst{1,2} \and
Qi Chu\inst{1,2} \and
Nenghai Yu\inst{1,2}}

\authorrunning{X.~Li et al.}

\institute{University of Science and Technology of China, China \and
Anhui Province Key Laboratory of Digital Security, China \and
The Chinese University of Hong Kong, China\\
\email{lxlkw@mail.ustc.edu.cn, yanlu@cuhk.edu.hk, flowice@ustc.edu.cn}\\
\email{\{jz\_li,qhyang233\}@mail.ustc.edu.cn, \{qchu,tgong,ynh\}@ustc.edu.cn}}

\maketitle

\begin{abstract}
\begin{sloppypar}
In person re-identification, neighbor-based methods have achieved significant success by interacting with neighbor samples to obtain more robust representations.
However, existing methods rely only on affinity relations, causing their success to depend heavily on the reliability of selected neighbors.
We find that affinity-only interaction often fails in challenging scenarios due to the inevitable presence of noisy neighbors.
To enable effective interactions under noisy neighborhoods, we revisit neighbor-based methods under distinct reliability conditions and propose a novel \textbf{Adaptive Neighbor Feature Interaction (ANFI)} method.
The core idea of ANFI is to account for negative effects from noisy neighbors, allowing samples to remain distinguishable from false positive neighbors.
Unlike existing methods, ANFI models not only affinity relations but also discrepancy relations, and employs sample-wise adaptive weighting for these two types of relations.
Given that capturing negative effects from noisy neighbors differs significantly from traditional relation learning, we derive discrepancy relations from a new \textbf{neighborhood similarity}, which provides more information than pairwise similarity.
In addition, we propose \textbf{Noisy Relation Supervision (NRS)} to train ANFI, gradually injecting robustness to noisy relations into the model.
Extensive experiments conducted under standard, cross-modal, and cross-domain settings, including comparisons with neighbor-based methods and re-ranking methods, demonstrate the superiority of our method across various neighbor distributions.

\end{sloppypar}
\end{abstract}
  
\section{Introduction}
\label{sec:intro}

\begin{figure}[!t]
    \centering
    \includegraphics[width=0.8\linewidth]{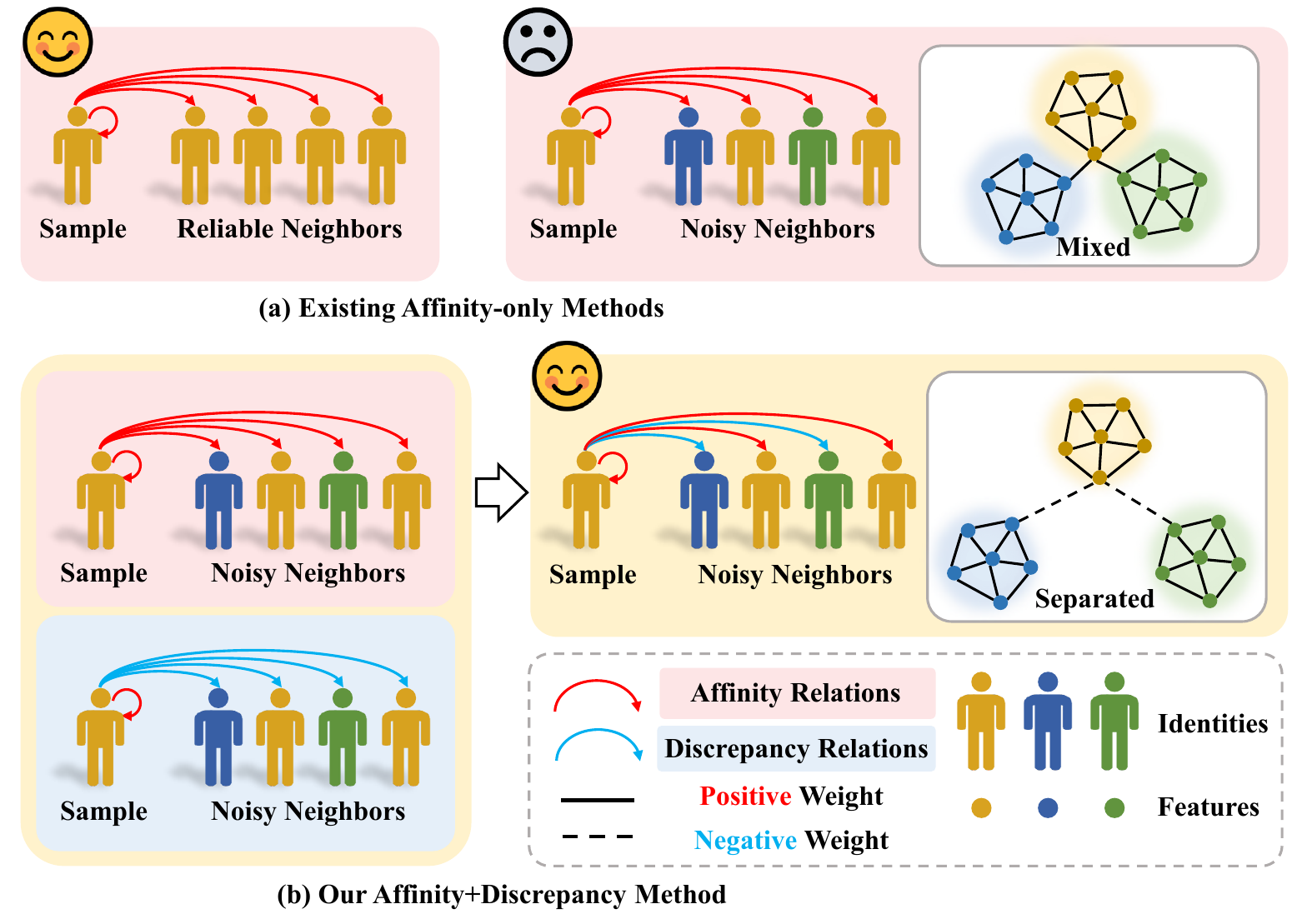}
    \caption{
    (a) Existing neighbor-based approaches always aggregate neighbor representations by affinity-only relations.
    (b) Our approach dynamically incorporates both positive (affinity) and negative (discrepancy) relations, allowing samples to remain distinguishable from false positive neighbors.
    }
    \label{fig:intro}
\end{figure}

Person re-identification (Re-ID) aims to retrieve a specific person from a gallery set containing a large number of images captured by different cameras and scenes.
Most research focuses on learning discriminative representations from single images~\cite{sun2018beyond,luo2019bag,ye2021deep,yang2023towards,li2024adaptive,li2025ATreid}.
However, due to the diversity of illumination, occlusion, viewpoint, and resolution, it is hard to extract sufficiently discriminative features from each image alone.

To overcome this limitation, neighbor-based methods~\cite{luo2019spectral,ye2020dynamic,lu2020cross,li2022counterfactual,wang2022nformer,zhang2023graph,yang2025cheb} have been developed to enhance image representations that contain not only their own information but also information from multiple related images.
In practice, these methods implement feature enhancement as a graph message-passing process.
Specifically, each image is seen as a node in a graph, with cross-image affinity represented as the edges of this graph. 
By applying graph convolution or transformer layers~\cite{wang2022nformer}, information from each image interacts with its neighbors, leading to richer and more discriminative image representations.

Unfortunately, the aforementioned scheme requires strict conditions to work well.
The calculated affinity has to be reliable to ensure that information from neighbors is related and complementary to the current image, rather than conflicting.
However, we observe that this condition does not always hold, especially in challenging situations, such as fuzzy query samples, small-scale gallery sets and cross-domain retrieval.
In these cases, affinities are not reliable, which makes the existing methods propagate irrelevant features between nodes.

To tackle this issue, we rethink neighbor relation learning under three distinct reliability conditions and conclude our main idea as follows:
1) In the good case of reliable neighbors, affinity relations effectively reduce the distance between related samples.
2) In the bad case of unreliable neighbors, affinity relations incorrectly reduce the distance between unrelated samples.
A new type of relation (discrepancy) is required to maintain the distinctions among these samples.
3) In the intermediate case of noisy neighbors, both affinity and discrepancy relations are necessary.

To achieve our idea, we propose a novel Adaptive Neighbor Feature Interaction (ANFI) model, which is a more comprehensive neighbor relation model that goes beyond affinity-only modeling.
As shown in \cref{fig:intro}, ANFI dynamically incorporates affinity (positive weight) and discrepancy (negative weight) relations, playing a role complementary to affinity and filtering out noisy propagation when faced with inaccurate neighbors.
In ANFI, we employ a new neighborhood similarity for discrepancy modeling, which takes into account neighborhood structural information beyond pairwise similarity, thereby enabling effective discrimination of noisy neighbors.
To train this new relational model, we propose a Noisy Relation Supervision (NRS) method that incorporates noise simulation and relational regularization.
The noise simulation effectively broadens the model's applicability beyond the low-noise neighbor distribution of the training set, while relational regularization supervises model outputs toward ground-truth relational features.

Moreover, we find that existing Re-ID testing setups may not fully cover diverse neighbor distributions.
A large number of simplistic gallery samples often lead to the dominance of low-noise neighbors, which can obscure the limitations of neighbor-based methods.
To provide a more comprehensive evaluation, we include testing settings that cover a wider range of neighbor distributions.
We compare ANFI with three categories of methods: pure image-based methods, neighbor-based methods, and re-ranking methods.
For a fair evaluation, we additionally report results under the same backbone to isolate the influence of backbone feature extraction.

Our main contributions are summarized as follows:

\noindent $\bullet$ We investigate the challenge of learning neighbor relations in person Re-ID, particularly in handling complex scenarios with noisy neighbor relations.
To address this, we propose a novel Adaptive Neighbor Feature Interaction (ANFI) model, which is the first Re-ID method to capture negative influences of noisy neighbors.

\noindent $\bullet$ To achieve ANFI, we design discrepancy relations with negative weights; introduce neighborhood similarity to incorporate structural information into relation modeling; and propose Noisy Relation Supervision (NRS) to enhance effective interaction among relevant features across various neighbor distributions.

\noindent $\bullet$ We conduct comprehensive evaluations of neighbor-based methods under various neighbor distributions. Extensive experiments on both standard and cross-domain settings demonstrate the superiority of our method over state-of-the-art methods.
 
\section{Related Work}
\label{sec:relate}

In this section, we first briefly review mainstream Re-ID methods that focus on learning single-image representations.
Then we introduce neighbor-based methods, which model sample interactions to improve representations.
Finally, we review ranking optimization methods that leverage neighbor information to refine ranking lists.

\noindent{\bf Person Re-ID.}
The key to person Re-ID is extracting discriminative features.
Most existing methods focus on learning better representations from single images.
Some methods design stronger feature extraction networks, while others introduce effective Re-ID loss functions to improve representation learning~\cite{sun2018beyond,ye2021deep,li2023clip,sun2020circle,li2026causal,li2026mfen}.
These methods fully exploit information from single images to learn discriminative features.

\noindent{\bf Neighbor-based Methods.}
Due to complex variations in Re-ID scenarios, such as changes in cameras, lighting, viewpoints, and occlusions, single-image feature learning methods often lead to unstable matching and sensitivity to outliers.
To address this challenge, neighbor-based methods conduct interactions among neighboring images in the representation space to obtain more robust features.
For instance, some methods use graph neural networks, while NFormer uses a transformer to extract additional information from k-nearest neighbors~\cite{luo2019spectral,zhang2020understanding,zhang2023graph,yang2025cheb,wang2022nformer}.
Furthermore, in visible-infrared Re-ID settings, many methods~\cite{ye2020dynamic,lu2020cross,li2022counterfactual} explore cross-modal pairwise relation learning with graph networks to extract complementary information from neighbors and alleviate modality discrepancies.
These methods encourage the model to cluster neighbor representations tightly, but the aggregation operation may still be constrained by false positive neighbors.
In contrast, our approach enables the model to adaptively learn both similarity and discrepancy, aiming to reduce the negative impact of false positive neighbors while preserving gains from true positive neighbors.

\noindent{\bf Ranking Optimization Methods.}
Ranking optimization is a post-processing technique used to improve the original ranking order at test time.
Recently, some approaches~\cite{ye2016person,sarfraz2018pose,zhong2017re,fang2023visible,chen2024jaccard} focus on exploring neighborhood similarity from the original ranking list itself for re-ranking, with the core idea that similar samples tend to share similar neighbors.
These methods operate directly on the initial ranking list and do not involve feature propagation.
In contrast, our method leverages representation-level interactions to yield more discriminative features.
 
\section{Revisiting Neighbor-based Methods and Analysis}
In this section, we first review neighbor-based methods. 
Next, we analyze the reasons for their success in scenarios with reliable neighbor distributions. 
Finally, we propose a solution to deal with noisy neighbors.

\subsection{Revisit of Neighbor-based Methods}
\begin{figure}[!t]
    \centering
    \includegraphics[width=0.7\linewidth]{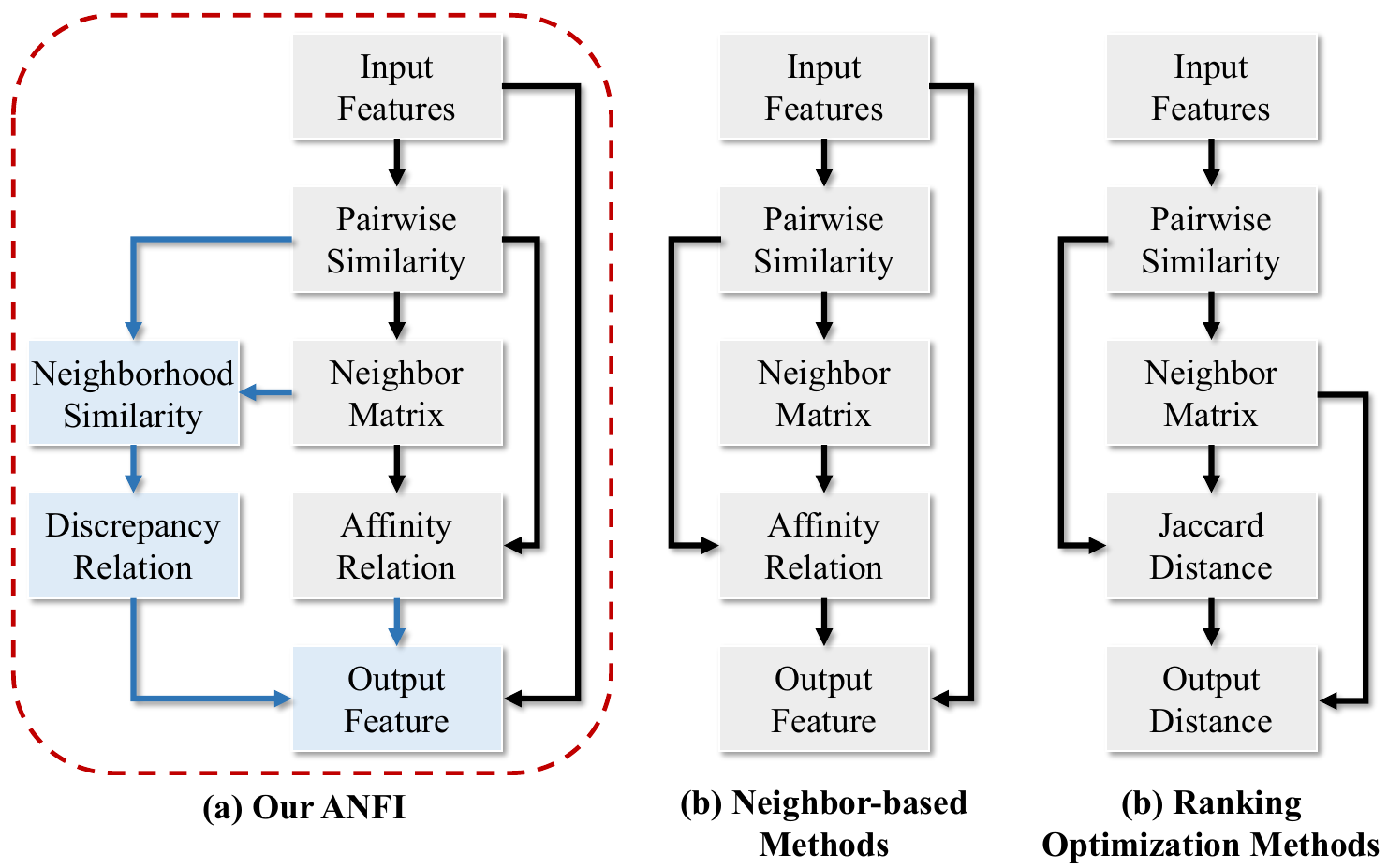}
    \caption{
    Pipelines of ANFI and related methods.
    }
    \label{fig:rethink}
\end{figure}

As shown in \cref{fig:rethink}, the pipeline of existing neighbor-based Re-ID methods can be summarized as follows:

\noindent $\bullet$ \textbf{Step 1: Input Feature $X$.} 
The backbone network extracts features $X=[x_1, ...,x_n]$ for all input images. Each image is considered a node in the graph, with $X$ as node features.

\noindent $\bullet$ \textbf{Step 2: Pairwise Similarity $S$.} 
$S\in\mathbb{R}^{n\times n}$ is derived from cosine similarity~\cite{luo2019spectral,li2022counterfactual}:
\begin{equation}  
\begin{aligned} 
S_{ij}=\exp(\cos(\varphi(x_i),\varphi(x_j))/\tau),
\label{eq:s}
\end{aligned}  
\end{equation}
where $\varphi$ is a projection function and $\tau$ is a temperature parameter.
Other methods compute $S$ by Euclidean distance~\cite{zhang2023graph,yang2025cheb} ($S_{ij}=\exp(-{||x_i-x_j||}_2/\tau)$) or dot product~\cite{wang2022nformer} ($S_{ij}=\exp(\varphi_q(x_i) \cdot \varphi_k(x_j)/\tau)$).

\noindent $\bullet$ \textbf{Step 3: Neighbor Matrix $N$.}
The $k$-nearest neighbors of the $i$-th sample are $\mathcal{N}_i=\{x_j|N_{ij}=1\}$. The neighbor matrix $N\in\{0,1\}^{n\times n}$ is computed
as follows:
\begin{equation}   
\begin{aligned}  
N=\mathrm{Topk}(S,k),
\end{aligned}   
\end{equation}
which sets the largest-$k$ values in each row of $S$ to 1, and sets the others to 0.
In this definition, the sample itself is also included in the neighbor set $x_i\in\mathcal{N}_i$.
$k$ is a critical hyper-parameter that is kept constant for all samples, but varies depending on the dataset and the testing setups.

Some methods further refine $N$ using reciprocal-nearest-neighbor filtering~\cite{wang2022nformer,zhang2023graph,yang2025cheb}: $\hat{N}_{ij}=N_{ij} N_{ji}$ or $(N_{ij}+N_{ji})/2$.

\noindent $\bullet$ \textbf{Step 4: Affinity Relations $\hat{A}$.} 
$\hat{A}\in\mathbb{R}^{n\times n}$ is calculated by combining the weights from $S$ with the connectivity from $N$, followed by row normalization:
\begin{equation}   
\begin{aligned}  
A=S\odot N,\quad
\hat{A}=D^{-1}A,
\label{eq:a=dad}
\end{aligned}   
\end{equation}
where $D\in\mathbb{R}^{n \times n}$ is the diagonal degree matrix for normalization, with $D_{ii}=\sum_jA_{ij}$.

\noindent $\bullet$ \textbf{Step 5: Affinity Features.} 
The enhanced features $F=[f_1,...,f_n]$ are obtained via message passing:
\begin{equation}   
\begin{aligned}  
F=\hat{A}\varphi(X),
\end{aligned}
\label{eq:f=ax}
\end{equation}
where $\varphi$ is a projection function.
Since the matrix $\hat{A}$ inherits the sparsity of $N$, sample interaction is constrained within a neighborhood, which filters out the influence of numerous irrelevant samples.

\noindent $\bullet$ \textbf{Step 6: Relation Learning.} 
The enhanced features $F$ are typically optimized with the identity loss:
\begin{equation}   
\begin{aligned}  
L_{id} = -\sum\nolimits_i y_i \log(P_i),
\end{aligned}   
\label{eq:idloss}
\end{equation}
where $y_i$ is the identity label, $P_i=\mathrm{Cls}(f_i)$ is the classification probability output by the classifier $\mathrm{Cls}(\cdot)$.

\subsection{Analysis of Neighbor-based Methods}
To further analyze neighbor-based Re-ID methods, we rewrite their core formula as follows:
\begin{equation}
\begin{aligned}
f_i=\sum\limits_{x_j\in\mathcal{N}_i^+}\hat{A}_{ij}\varphi(x_j)+\sum\limits_{x_j\in\mathcal{N}_i^-}\hat{A}_{ij}\varphi(x_j),\quad
\sum\limits_{x_j\in\mathcal{N}_i}\hat{A}_{ij}=1,
\end{aligned}   
\end{equation}
where $\hat{A}_{ij}\ge0$, $\mathcal{N}_i^+=\{x_j|x_j\in\mathcal{N}_i,y_i= y_j\}$ is the positive neighbor set, $\mathcal{N}_i^-=\{x_j|x_j\in\mathcal{N}_i,y_i\ne y_j\}$ is the negative neighbor set.

\noindent $\bullet$ \textbf{Fully Reliable Neighbors (Good Case).} 

When all neighbors are positive samples (i.e., $|\mathcal{N}_i^-|=0$), the affinity relations assign positive weights to positive samples, effectively reducing the distance between positive samples.

\noindent $\bullet$ \textbf{Fully Unreliable Neighbors (Bad Case).} 

When all non-self neighbors are negative samples (i.e., $|\mathcal{N}_i^+|=1$ and $|\mathcal{N}_i^-|=k-1$), the affinity relations assign positive weights to negative samples, mistakenly decreasing the distance between positive and negative samples.

To ensure that the neighbor relations are suitable for this case, we need to assign negative weights to all non-self neighbors.
The relational model for the bad case can be modified as follows:
\begin{equation}
\begin{aligned}
f_i^d=\varphi'(x_i)+\sum\limits_{x_j\in\mathcal{N}_i}(-\hat{A}_{ij}^d)\varphi'(x_j),\quad
\sum\limits_{x_j\in\mathcal{N}_i}\hat{A}_{ij}^d=1,
\label{eq:f=(i-a)x}
\end{aligned}   
\end{equation}
where $\hat{A}^d$ is the discrepancy relation matrix.
The weight assigned to itself is $(1-\hat{A}^d_{ii})>0$ and the weights assigned to other neighbors are $(-\hat{A}^d_{ij})<0$.
\cref{eq:f=(i-a)x} can be rewritten as follows:  
\begin{equation}
\begin{aligned}
f_i^d=\sum\limits_{j\ne i,x_j\in\mathcal{N}_i}\hat{A}_{ij}^d(\varphi'(x_i)-\varphi'(x_j)).
\label{eq:f=a(x-x)}
\end{aligned}   
\end{equation}
It is clear from \cref{eq:f=a(x-x)} that it captures the discrepancy between samples and their neighbors; thus $\hat{A}^d$ reflects discrepancy relations.

\noindent $\bullet$ \textbf{Noise Neighbors (Intermediate Case).}

In practice, neighbors contain both positive and negative samples. Therefore, it is reasonable to simultaneously consider both the affinity relations $\hat{A}$ and the discrepancy relations $\hat{A}^d$.
For a sample $x_i$, a lower $\mathrm{nnr}(x_i)$ implies stronger reliance on affinity interaction, whereas a higher $\mathrm{nnr}(x_i)$ calls for stronger discrepancy interaction.

To quantify this trade-off, we define the Noise Neighbor Ratio (NNR) as a sample-level noise indicator:
\begin{equation}        
\begin{aligned}   
 \mathrm{nnr}(x_i)=\frac{|\mathcal{N}_i^-|}{|\mathcal{N}_i|},\quad \mathrm{nnr}(X) = \frac{1}{n}\sum_i \mathrm{nnr}(x_i).
\label{eq:nni}
\end{aligned}       
\end{equation}
Therefore, $\mathrm{nnr}(x_i)$ specifies the desired mixing trend between $f_i$ and $f_i^d$, while the final sample-wise mixing is learned by the model.

\section{Proposed Method}  
\label{sec:method}
\begin{figure*}[!t]
    \centering
    \includegraphics[width=1.0\linewidth]{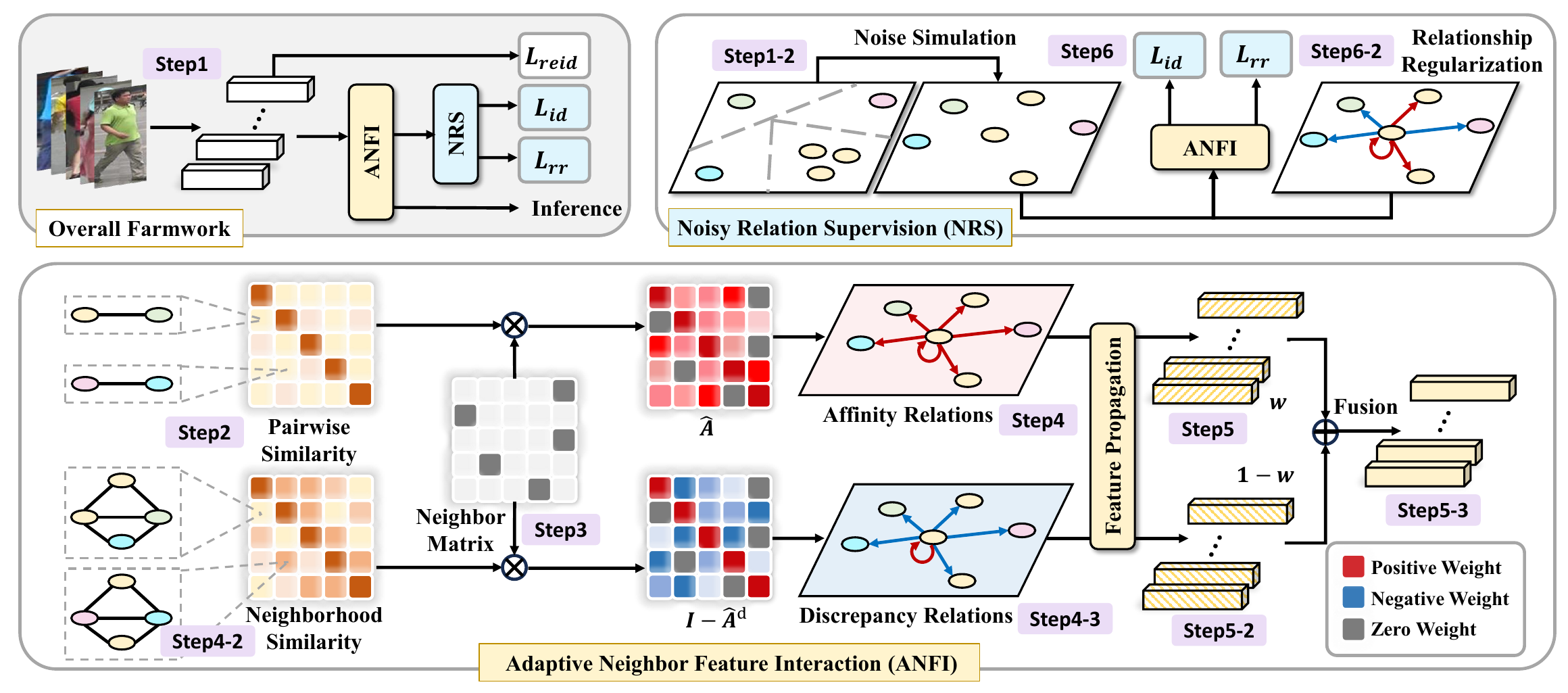}
    \caption{
    The framework of the proposed Adaptive Neighbor Feature Interaction (ANFI) model achieves robust representations through interactions among multiple images.
    }
    \label{fig:method}
\end{figure*}

As shown in \cref{fig:method}, we introduce Adaptive Neighbor Feature Interaction (ANFI) and Noisy Relation Supervision (NRS) at both the model and training levels.

\subsection{Adaptive Neighbor Feature Interaction}
Based on the previous revisit and analysis, the existing methods establish affinity relations (Step 4, \cref{eq:a=dad}) and utilize the established relations for feature interaction (Step 5, \cref{eq:f=ax}).
This affinity is applicable only to reliable neighbors, and when faced with noisy neighbors, it is necessary to consider discrepancy relations.

\noindent $\bullet$ \textbf{Step 4-2: Neighborhood Similarity $E'$.}
Existing methods derive sample relations through pairwise similarity $S$, which is more suitable for affinity relations rather than for complex discrepancy relations.
Therefore, in Step 4, we introduce neighborhood similarity $E'\in\mathbb{R}^{n\times n}$ to explore the differences between samples by leveraging multiple third-party samples $x_k$.
$E'$ can be calculated as follows:
\begin{equation}       
\begin{aligned}  
E'_{ij} = \sum_{x_k\in\mathcal{N}_i\cap\mathcal{N}_j} S'_{ik}\cdot S'_{jk}=\sum_{k} S'_{ik}\cdot S'_{jk}\cdot N_{ik}\cdot N_{jk},
\end{aligned}       
\end{equation}
Here, third-party samples are automatically selected as shared neighbors $x_k\in\mathcal{N}_i\cap\mathcal{N}_j$; their number is determined by neighborhood overlap under $k$, without introducing additional hyper-parameters.
where $S_{ij}'=\exp(\cos(\varphi'(x_i),\varphi'(x_j))/\tau)$ is computed with a distinct projection function $\varphi'$.
$E'$ considers information from the shared neighborhoods $\mathcal{N}_i \cap \mathcal{N}_j$ of two samples, providing neighborhood structure information beyond pairwise similarity.
This approach differs from existing reciprocal nearest neighbors~\cite{wang2022nformer,zhang2023graph}, which do not consider third-party samples.

\noindent $\bullet$ \textbf{Step 4-3: Discrepancy Relations $\hat{A}^d$.}
After obtaining the neighborhood similarity, we model the discrepancy relations as follows:
\begin{equation}   
\begin{aligned}  
A^d=E'\odot N,\quad
\hat{A}^d=({D^d})^{-1}A^d.
\end{aligned}   
\end{equation}
where $N$ is the neighbor matrix, $D^d$ is a diagonal matrix for normalization and $D^d_{ii}=\sum_jA^d_{ij}$. 

\noindent $\bullet$ \textbf{Step 5-2: Discrepancy Features $F^d$.}
Based on the analysis in \cref{eq:f=(i-a)x}, we use $\hat{A}^d$ for feature interaction:
\begin{equation}   
\begin{aligned}  
F^d=(I-\hat{A}^d)\varphi'(X).
\end{aligned}
\end{equation}

\noindent $\bullet$ \textbf{Step 5-3: Adaptive Mixing.}
Finally, we dynamically calculate the sample-wise importance scores $w_i$ to effectively fuse $f_i$ and $f_i^{d}$ and obtain the final representation $f_i^{mix}$:
\begin{equation}     
\begin{aligned}
(1-w_i;w_i) = \mathrm{softmax}(h(f_i);h'(f_i^{d})),
\end{aligned}     
\end{equation}
\begin{equation}       
\begin{aligned}  
f_i^{mix} = (1-w_i)\cdot f_i+w_i\cdot f_i^{d},
\end{aligned}       
\end{equation}
where $h(\cdot)$ and $h'(\cdot)$ are two linear layers with an output dimension of 1.
For different samples, we obtain distinct values $w_i$ so that $f_i^{mix}$ can adaptively leverage the affinity and discrepancy relations among neighbors.

Compared with existing neighbor-based methods that focus solely on the positive influence of neighbors, ANFI simultaneously considers both affinity and discrepancy among neighbors.
ANFI also provides an adaptive weighting strategy that enables our method to apply to diverse neighbor distributions, ranging from fully reliable neighbors to fully unreliable neighbors.

\subsection{Noisy Relation Supervision}
\label{sec:nrcl}
\noindent $\bullet$ \textbf{Step 1-2: Noise Simulation.}
We observe that the graph constructed during training by existing methods is biased, exhibiting a low Noise Neighbor Ratio (NNR).
This phenomenon is attributed to the supervision of the backbone, which leads to high similarity among intra-identity features and low similarity among inter-identity features in the training set, ultimately resulting in a predominance of reliable neighbor samples.
At test time, unseen identities typically come with noisy neighbors, making it difficult for relation models trained on reliable neighbors to generalize.

To address this issue, we add noise to the image features $x_i$ during the training process to simulate a testing environment with a noisy neighbor distribution:
\begin{equation}        
\begin{aligned}
x_i := (1-\alpha_n)\cdot x_i+\alpha_n\cdot \mathrm{sg}(x_j),
\label{eq:alpha}
\end{aligned}       
\end{equation}
where $:=$ denotes feature update, $\mathrm{sg}(\cdot)$ denotes the stop-gradient operation, $x_j$ is a randomly sampled feature in the training batch, and $\alpha_n$ is sampled from a uniform distribution:
\begin{equation}        
\begin{aligned}
\alpha_n\sim\mathcal{U}(0, \alpha_{max}\cdot (1-\mathrm{nnr}(X))),
\label{eq:alpha_n}
\end{aligned}       
\end{equation}
where $\alpha_{max}$ represents the upper limit of noise intensity, and $\mathrm{nnr}(X)$ is the batch-level Noise Neighbor Ratio computed from ground-truth identity labels using \cref{eq:nni} before noise simulation.
Therefore, cleaner batches (lower $\mathrm{nnr}(X)$) receive stronger perturbation, while noisier batches receive milder perturbation; this adaptation is used only during training.
As training progresses, the noise intensity is adaptively adjusted, gradually injecting noise tolerance into the model.

\noindent $\bullet$ \textbf{Step 6-2: Relationship Regularization.}
Furthermore, the commonly used identity loss $L_{id}$ in existing methods only constrains the final output features of the graph to perform effective classification, without directly ensuring the accuracy of each neighbor relationship.

To learn neighbor relationships more accurately, we propose a new relationship regularization.
Specifically, we set the weights of negative neighbors in the affinity matrix $\hat{A}$ to zero to obtain the ground truth affinity feature $g_i$. 
Similarly, we set the weights of positive neighbors (except the sample itself) in the discrepancy matrix $\hat{A}^d$ to zero to obtain the ground truth discrepancy feature $g_i^d$. 
After masking, we re-normalize each row of the two relation matrices so that the remaining weights sum to 1 before computing $g_i$ and $g_i^d$.
Finally, we replace the adaptive weights $w_i$ with $\mathrm{nnr}(x_i)$ for each sample, yielding the final ground truth feature $g^{mix}_i$:
\begin{equation}
\begin{aligned}
    g^{mix}_i = (1-\mathrm{nnr}(x_i))\cdot g_i+\mathrm{nnr}(x_i) \cdot g_i^d.
\label{eq:gt}
\end{aligned}
\end{equation}
This replacement is only used to construct supervision targets; during inference, the sample-wise mixing weights are still predicted by the adaptive mixing module.
By matching $f_i^{mix}$ to $g_i^{mix}$ through $L_{rr}$, this supervision propagates the $\mathrm{nnr}(x_i)$ trend to $w_i$, encouraging larger $w_i$ under noisier neighbors.

Next, we propose a relationship regularization loss to constrain the learned graph feature to be close to the ground-truth feature:
\begin{equation}
\begin{aligned}
    \mathcal{L}_{rr} = \mathbb{E}[\mathrm{D_{KL}}(\mathrm{Cls}({f_i^{mix}})||\mathrm{Cls}(g_i^{mix}))],
\label{eq:kl}
\end{aligned}
\end{equation}
where $\mathrm{D_{KL}}(\cdot||\cdot)$ denotes the KL divergence, $\mathrm{Cls}(\cdot)$ represents the classifier in \cref{eq:idloss}.
This regularization effectively constrains the relational model to distinguish affinity and discrepancy relations.

The whole model is trained end-to-end, and the overall loss comprises the Re-ID loss~\cite{luo2019bag,li2022counterfactual} for training the backbone feature extraction network, as well as the identity loss $L_{id}$ in \cref{eq:idloss} and the relationship regularization loss $L_{rr}$ for training our ANFI module:
\begin{equation}
\begin{aligned}
\mathcal{L}=\mathcal{L}_{reid}(x_i)+\mathcal{L}_{id}(f_i^{mix})+\mathcal{L}_{rr}(f_i^{mix},g_i^{mix}).
\label{eq:loss}
\end{aligned}
\end{equation}
 
\section{Experiments}  
\label{sec:exper}

\noindent{\bf Datasets.}
We conduct comprehensive experiments to evaluate our method on five benchmarks.
These include three standard Re-ID datasets, Market1501~\cite{zheng2015scalable}, CUHK03~\cite{li2014deepreid}, and MSMT17~\cite{wei2018person}, as well as two visible-infrared cross-modal Re-ID datasets, SYSU-MM01~\cite{wu2017rgb} and RegDB~\cite{nguyen2017person}.
Market1501 contains 1,501 identities with 32,217 images captured by 6 cameras.
CUHK03 contains 1,467 identities with 14,097 images captured by 2 cameras.
MSMT17 contains 4,101 identities with 126,441 images captured by 15 cameras.
SYSU-MM01 contains 491 identities with 29,033 visible images and 15,712 infrared images captured by 4/2 RGB/IR cameras.
RegDB contains 412 identities with 4,120 visible images and 4,120 infrared images captured by 1/1 RGB/IR cameras.
Across Market1501, CUHK03, MSMT17, SYSU-MM01, and RegDB, each identity contains an average of 21, 7.6, 27, 3.1, and 10 gallery images, respectively.
These statistics are used as coarse references when setting the search range of neighborhood size $k$.

\noindent{\bf Evaluation Protocol.}
The experiments follow existing standard evaluation settings~\cite{wang2022nformer,yang2025cheb} and employ two assessment metrics: Cumulative Matching Characteristic (CMC) and mean Average Precision (mAP).
Following~\cite{wang2022nformer,yang2025cheb}, we eliminate interactions among different query images (single-query setup).

\noindent{\bf Implementation Details}
Our method is implemented using the PyTorch framework.
We employ ResNet-50~\cite{he2016deep} as our backbone network and draw upon improvements from the widely used baseline model AGW~\cite{ye2021deep}.
The images are resized to a fixed resolution of 384$\times$192.
We adopt random cropping, random horizontal flipping, and random erasing~\cite{zhong2020random} for data augmentation.
For cross-modal Re-ID, we also incorporate random channel enhancement~\cite{ye2021channel}, and the loss term $L_{reid}$ in \cref{eq:loss} is derived from the baseline provided by CIFT~\cite{li2022counterfactual}.
We use the SGD optimizer to train the model for 120 epochs.
The learning rate is set to 0.01 with a warm-up scheme and cosine decay schedule.
The batch size is set to 64 with 8 identities per mini-batch.
We set the projection function $\varphi$ and $\varphi'$ to batch normalization layers.
The temperature parameter $\tau$ is set to 0.4, while the noise parameter $\alpha_{max}$ is set to 0.4.

\subsection{Comparison with SOTA methods}

\begin{table}[!t]
    \fontsize{8}{8}\selectfont
    \setlength\tabcolsep{1.3pt}
    \renewcommand{\arraystretch}{1.3}
    \centering
    \caption{Comparison of rank-1 (R1) and mAP (\%) with representative SOTA methods on Market1501, CUHK03, MSMT17, SYSU-MM01 and RegDB datasets.}
    \begin{tabular}{l c c c c c c @{\hspace{4pt}}|@{\hspace{4pt}} l c c c c}
        \toprule[1pt]
        \multirow{2}{*}{Method} &
        \multicolumn{2}{c}{Market1501} &
        \multicolumn{2}{c}{CUHK03} &
        \multicolumn{2}{c@{\hspace{4pt}}|@{\hspace{4pt}}}{MSMT17} &
        \multirow{2}{*}{Method} &
        \multicolumn{2}{c}{SYSU-MM01} &
        \multicolumn{2}{c}{RegDB}
        \cr
        \cmidrule(r){2-7}\cmidrule(l){9-12}
        & mAP & R1 & mAP & R1 & mAP & R1 &
        & mAP & R1 & mAP & R1
        \cr\midrule
        CAL~\cite{rao2021counterfactual}&
        89.5&95.5&-&-&64.0&84.2&
        CAJ~\cite{ye2021channel}&
        66.9&69.9&78.5&84.9
        \cr
        CLIP-ReID~\cite{li2023clip}&
        89.6&95.5&81.6&80.9&73.4&88.7&
        MPANet~\cite{wu2021discover}&
        68.2&70.6&80.8&83.3
        \cr
        Instruct-ReID~\cite{he2024instruct}&
        93.5&96.5&85.4&86.5&72.4&86.9&
        DEEN~\cite{zhang2023diverse}&
        71.8&74.7&84.3&90.3
        \cr
        VersReID~\cite{zheng2024versatile}&
        93.2&96.8&-&-&74.2&88.8&
        PartMix~\cite{kim2023partmix}&
        74.6&77.8&82.4&85.3
        \cr
        CA-Jaccard~\cite{chen2024jaccard}&
        94.5&96.2&-&-&74.1&86.2&
        DNS~\cite{jiang2024domain}&
        74.4&77.3&88.4&93.3
        \cr
        NFormer~\cite{wang2022nformer}&
        93.0&95.7&76.4&79.0&62.2&80.8&
        SAAI~\cite{fang2023visible}&
        77.0&75.9&91.8&91.6
        \cr
        GCR~\cite{zhang2023graph}&
        95.1&96.6&89.0&89.7&76.9&86.8&
        cm-SSFT~\cite{lu2020cross}&
        54.1&47.7&64.9&64.6
        \cr
        Cheb-GR~\cite{yang2025cheb}&
        93.2&95.2&85.4&83.9&-&-&
        CIFT~\cite{li2022counterfactual}&
        74.8&74.1&91.4&91.2
        \cr
        ANFI (Ours)&
        \textbf{95.7}&\textbf{96.8}&\textbf{90.7}&\textbf{91.3}&\textbf{82.9}&\textbf{90.3}&
        ANFI (Ours)&
        \textbf{81.1}&\textbf{80.3}&\textbf{94.4}&\textbf{93.7}
        \cr\bottomrule[1pt]
    \end{tabular}
    \label{tab:comp_sota}
\end{table}

We compare ANFI with representative state-of-the-art (SOTA) methods on Market1501, CUHK03, MSMT17, SYSU-MM01, and RegDB in \cref{tab:comp_sota}.

\noindent \textbf{Results on Standard Re-ID Datasets.}
As shown in the left block of \cref{tab:comp_sota}, ANFI achieves the best average mAP and rank-1 accuracy across the three standard Re-ID datasets.
Specifically, the mAP of ANFI surpasses that of the single-image representation method Instruct-ReID~\cite{he2024instruct} by 2.2\% / 5.3\% / 10.5\% on Market1501 / CUHK03 / MSMT17, respectively.
The gain is larger on mAP than on rank-1, indicating that ANFI improves overall ranking quality rather than only top retrieval.

ANFI also outperforms the neighbor-based methods Cheb-GR~\cite{yang2025cheb} and GCR~\cite{zhang2023graph} in terms of mAP on these datasets.
Compared with the re-ranking method CA-Jaccard~\cite{chen2024jaccard}, ANFI also achieves better mAP.

\noindent \textbf{Results on Cross-modal Re-ID Datasets.}
As shown in the right block of \cref{tab:comp_sota}, ANFI outperforms the neighbor-based methods cm-SSFT~\cite{lu2020cross} and CIFT~\cite{li2022counterfactual}, as well as the re-ranking method SAAI~\cite{fang2023visible}.
Specifically, on SYSU-MM01 / RegDB, ANFI improves mAP by 4.1\% / 2.6\% over SAAI~\cite{fang2023visible}.
Compared with CIFT~\cite{li2022counterfactual}, the gains are 6.3\% / 3.0\%, respectively.
Given the large modality gap in cross-modal retrieval, these gains indicate stronger robustness under noisier neighbor distributions.

\subsection{Comparison Under the Same Backbone}
To isolate the effect of backbone feature extraction and provide a fair apples-to-apples comparison among gallery-based methods, we further report comparisons under the same backbone.
For each dataset-backbone pair, we compare K-reciprocal~\cite{zhong2017re}, CA-Jaccard~\cite{chen2024jaccard}, Cheb-GR~\cite{yang2025cheb}, GCR~\cite{zhang2023graph}, and ANFI (ours) on top of the same backbone features while using gallery information at inference.
This protocol also verifies that ANFI can be flexibly plugged into different single-image backbones.

\begin{table}[!t]
    \fontsize{8}{8}\selectfont
    \setlength\tabcolsep{1.4pt}
    \renewcommand{\arraystretch}{1.3}
    \centering
    \caption{Comparison under the same backbone. For fair evaluation, we report mAP (\%) for backbone-only features (`Raw') and backbone + gallery-based methods.}
    \begin{tabular}{l l c c c c c c}
        \toprule[1pt]
        Dataset & Backbone & Raw & K-recip.~\cite{zhong2017re} & CA-Jacc.~\cite{chen2024jaccard} & Cheb-GR~\cite{yang2025cheb} & GCR~\cite{zhang2023graph} & ANFI
        \cr\midrule
        CUHK03 & TransReID~\cite{he2021transreid} & 75.4 & 86.4 & 86.9 & 85.4 & 84.0 & \textbf{88.1}
        \cr
        MSMT17 & CLIP-ReID~\cite{li2023clip} & 73.4 & 84.1 & 86.0 & 83.3 & 83.5 & \textbf{86.5}
        \cr
        SYSU-MM01 & DEEN~\cite{zhang2023diverse} & 71.8 & 77.2 & 77.2 & 75.8 & 76.3 & \textbf{79.3}
        \cr\bottomrule[1pt]
    \end{tabular}
    \label{tab:fair_backbone}
\end{table}

As shown in \cref{tab:fair_backbone}, ANFI achieves the best mAP on all datasets.
Compared with the strongest baseline in each row, ANFI further improves mAP by 1.2\% / 0.5\% / 2.1\% on CUHK03 / MSMT17 / SYSU-MM01, respectively.
We attribute this to the fact that K-reciprocal~\cite{zhong2017re}, CA-Jaccard~\cite{chen2024jaccard}, Cheb-GR~\cite{yang2025cheb}, and GCR~\cite{zhang2023graph} mainly rely on affinity-based neighbor interaction, which is more sensitive to noisy neighbors, while ANFI jointly models affinity and discrepancy relations for more robust gallery-based interaction.

\subsection{Comprehensive Testing Benchmark}
\begin{table*}[!t]
    \fontsize{8}{8}\selectfont
    \setlength\tabcolsep{5.0pt}
    \renewcommand{\arraystretch}{1.2}
    \centering
    \caption{
    The $\Delta$ mAP performance of Re-ID methods on different neighbor distributions. `n-shot' means that the average number of gallery images per identity is $n$. `M$\rightarrow$C' and `M$\rightarrow$S' denote cross-domain testing from Market1501 to CUHK03 and from Market1501 to SYSU-MM01, respectively.
    }
    \begin{tabular}{l c c c c c c c}
        \toprule[1pt]
        &         
        \multicolumn{5}{c}{Market1501}&         
        \multicolumn{2}{c}{Cross-Domain}
        \cr\cmidrule(r){2-6}\cmidrule(r){7-8}
        \multirow{-2}{*}{Method}&
        {Standard}&
        {1-shot}&
        {2-shot}&
        {Hard Query}&
        {Poor Model}&
        {M$\rightarrow$C}&
        {M$\rightarrow$S}
        \cr\midrule
        \rowcolor[HTML]{EFEFEF}\multicolumn{8}{c}{Ranking Optimization Methods}\cr
        K-recip.~\cite{zhong2017re}&
        +4.8&-0.1&+1.6&-1.5&+0.8&+2.8&-2.2
        \cr         
        ECN~\cite{sarfraz2018pose}&
        +4.5&-1.5&+1.4&-2.1&-0.2&+2.2&-3.0
        \cr         
        SAAI~\cite{fang2023visible}&
        +4.0&-0.9&+1.5&-0.8&+0.6&+2.4&-1.9
        \cr         
        CA-Jacc.~\cite{chen2024jaccard}&
        +4.9&-0.3&+1.0&-1.9&+0.9&+2.7&-2.2
        \cr\midrule
        \rowcolor[HTML]{EFEFEF}\multicolumn{8}{c}{Neighbor-based Methods}\cr
        SFT~\cite{luo2019spectral}&
        +1.8&+0.1&+0.3&-0.3&+0.3&+1.9&-0.8
        \cr
        NFormer~\cite{wang2022nformer}&
        +1.5&-0.8&-0.1&-1.8&+0.2&+1.3&-2.5
        \cr
        GCR~\cite{zhang2023graph}&
        +4.7&-1.0&+1.4&-1.4&+0.5&+1.9&-1.6
        \cr
        Cheb-GR~\cite{yang2025cheb}&
        +3.8&-0.7&+1.1&-1.0&+0.5&+1.6&-1.9
        \cr\midrule
        ANFI (Ours)&
        \textbf{+5.4}&
        \textbf{+0.9}&
        \textbf{+2.3}&
        \textbf{+0.6}& 
        \textbf{+1.1}&
        \textbf{+3.5}&
        \textbf{+0.2}
        \cr\bottomrule[1pt]
    \end{tabular}
    \label{tab:comp3}
\end{table*}

To facilitate a comprehensive and fair evaluation of neighbor-based methods and ranking optimization methods, we conduct an in-depth investigation under varying neighbor distributions, as shown in \cref{tab:comp3}.
The benchmark settings are as follows:

\noindent 1) 
In common Re-ID tests, reliable neighbors dominate, which can obscure the inadequacies of affinity-only relations.
Following recent neighbor-based methods~\cite{yang2025cheb,wang2022nformer}, we conduct \textbf{cross-domain} testing and \textbf{small-scale gallery} testing to encompass a broader range of neighbor cases, particularly noisy neighbor distributions.
For small-scale gallery testing, we sample several gallery subsets, each containing $n$ images per identity, referred to as `$n$-shot' testing.
Additionally, we conduct tests involving `Hard Query' and `Poor Model'. `Hard Query' denotes the $10\%$ query samples with the lowest Rank-1 accuracy under the baseline model; this subset is selected using the baseline only, avoiding circular evaluation. `Poor Model' is simulated by prematurely terminating model training (30/120 epochs), thereby representing a weaker baseline.

\noindent 2) Neighbor-based methods are plug-and-play modules, while ranking optimization methods are training-free. To ensure a fair comparison, we use the same baseline model for all approaches to extract single-image representations and report the performance variations ($\Delta$ mAP) relative to this baseline.

\noindent 3) Due to substantial differences in neighbor distributions across testing modes, we adjust neighbor-related hyper-parameters for all methods in each configuration, e.g., $k$ in ANFI, $k_1$ and $k_2$ in K-reciprocal~\cite{zhong2017re}, and $\lambda$ in Cheb-GR~\cite{yang2025cheb}.

Based on the experimental results, the following conclusions can be drawn:

\noindent 1) When the number of images per identity in the test set is low, such as 1 or 2, a significant number of false positive neighbors may arise. 
Conducting relation learning in such a noisy environment is extremely challenging, and most existing methods tend to reduce retrieval performance. 
This reveals the shortcomings of relying solely on affinity relations.

\noindent 2) 
Moreover, challenging query samples can lead to significantly noisier neighbors. Because this subset is relatively small, its impact is difficult to reflect in standard tests. Compared with same-task baseline degradation (`Poor Model'), cross-domain transfer shows asymmetric behavior: `M$\rightarrow$S' causes clear performance drops for most methods, while `M$\rightarrow$C' usually still yields gains but with limited margins.

\noindent 3) Our ANFI method effectively handles noisy relations and achieves the largest gains across all seven settings.
ANFI is the only method with positive $\Delta$ mAP in all seven settings (+5.4/+0.9/+2.3/+0.6/+1.1/+3.5/+0.2), while other methods show negative $\Delta$ mAP in 2--4 settings.
SFT~\cite{luo2019spectral} remains positive in 5/7 settings, but its gains are smaller than ANFI.
This indicates that discrepancy relation modeling complements affinity relations and improves robustness under noisy neighbors.

\noindent  4) 
Moreover, our ANFI method remains effective under reliable relations, achieving a 5.4\% improvement in mAP that surpasses other methods in the standard test mode on Market1501.

\begin{table}[t]
    \renewcommand{\arraystretch}{1.2}      
    \fontsize{8}{8}\selectfont
    \setlength\tabcolsep{4.0pt}
    \centering
    \caption{
    Ablation study on each component of ANFI on Market1501 under `Standard', `1-shot', and `Hard Query' test modes.
    mAP accuracy (\%) is reported.
    The terms `Affi' and `Disc' refer to affinity and discrepancy relations.
    }
    \begin{tabular}{c cccc cc c}
        \toprule[1pt]
        &
        \multicolumn{2}{c}{ANFI}&
        \multicolumn{2}{c}{NRS}&
        &
        \cr\cmidrule(r){2-5}
        \multirow{-2}{*}{i}&
        {Affi.}&{Disc.}&
        {Noise}&{$L_{rr}$}&
        \multirow{-2}{*}{Standard}&
        \multirow{-2}{*}{1-shot}&
        \multirow{-2}{*}{Hard Query}
        \cr\midrule
        1&&&&&
        90.3&93.5&65.5
        \cr
        2&\Checkmark&&&&
        94.2&92.5&63.1
        \cr
        3&\Checkmark&\Checkmark&&&
        94.8&93.5&65.3
        \cr   
        4&\Checkmark&\Checkmark&\Checkmark&&
        95.1&93.8&65.9
        \cr
        5&\Checkmark&\Checkmark&\Checkmark&
        \Checkmark&
        95.7&94.4&66.1
        \cr
        \bottomrule[1pt]
    \end{tabular}
    \label{tab:abl1}
\end{table}

\subsection{Ablation Study}

To evaluate the contribution of each component in ANFI, we conduct ablation experiments on Market1501 under `Standard', `1-shot', and `Hard Query' test modes, and report mAP in \cref{tab:abl1}.

\noindent {\bf Effectiveness of ANFI.} 
As shown in the 2$^{nd}$ row, applying affinity relations improves mAP by 3.9\%.
This indicates that affinity relations among reliable neighbors can outperform single-image representations.
However, under the `1-shot' and `Hard Query' test modes, noisy neighbors cause mAP drops of 1.0\% and 2.4\%, respectively.
As illustrated in the 3$^{rd}$ row, introducing discrepancy relations enhances robustness across different neighbor distributions.
Compared with using only affinity relations, mAP increases by 1.0\% and 2.2\% under the `1-shot' and `Hard Query' test modes.

\noindent {\bf Effectiveness of NRS.} 
As shown in the 4$^{th}$ and 5$^{th}$ rows, we analyze the effects of relationship regularization $L_{rr}$ and noise simulation in NRS.
Experimental results indicate that both components facilitate noisy-relation learning and improve retrieval performance, with total gains of 0.9\%, 0.9\%, and 0.8\% under the `Standard', `1-shot', and `Hard Query' modes.
With NRS, the model reaches the best result, i.e., 95.7\% mAP on Market1501.
Most importantly, with ANFI and NRS, the model still achieves improvements in challenging scenarios where there is at most one non-self positive sample among neighbors.

These ablation experiments demonstrate that ANFI and NRS are complementary and together provide robust multi-image relational learning.

\subsection{Adaptive Weight Analysis}

\begin{figure}[!t]
    \centering
    \includegraphics[width=0.7\linewidth]{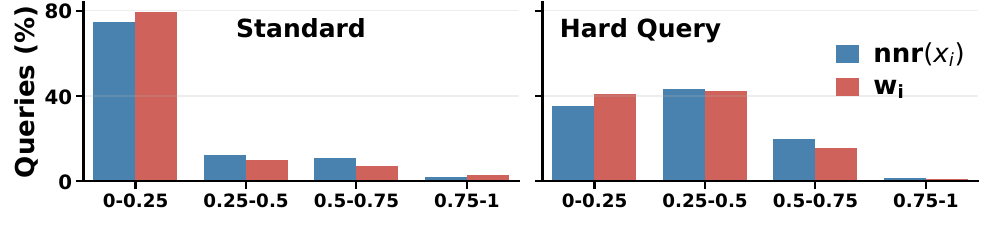}
    \caption{Distributions of oracle $\mathrm{nnr}(x_i)$ and learned adaptive weights $w_i$ on Market1501 under Standard and Hard Query settings.}
    \label{fig:nnr_wi}
    \vspace{-3mm}
\end{figure}

\cref{fig:nnr_wi} compares the learned adaptive weight $w_i$ with the oracle noise neighbor ratio $\mathrm{nnr}(x_i)$.
In the Standard setting, clean neighborhoods dominate, with 74.8\% of samples having $\mathrm{nnr}(x_i)=0$.
In contrast, the Hard Query setting contains only 35.3\% clean neighborhoods, confirming a substantially noisier neighbor distribution.
With 0.25-width bins, the learned $w_i$ follows the trend of $\mathrm{nnr}(x_i)$ in both settings and does not collapse to discrepancy-dominant behavior.
Quantitatively, the mean absolute difference $\frac{1}{|\mathcal{Q}|}\sum_i|w_i-\mathrm{nnr}(x_i)|$ is 0.07 in the Standard setting and 0.12 in the Hard Query setting.
We further replace the learned $w_i$ with test-time oracle $\mathrm{nnr}(x_i)$, which decreases mAP by 0.1 and 0.3 in the Standard and Hard Query settings, respectively.
This indicates that the final-feature supervision encourages $w_i$ to follow the neighbor-noise trend while still preserving discriminative cues beyond the raw count of noisy neighbors.
\subsection{Efficiency Analysis}

\begin{table}[H]
    \renewcommand{\arraystretch}{1.2}
    \fontsize{8}{8}\selectfont
    \setlength\tabcolsep{2.0pt}
    \centering
    \caption{Practical end-to-end testing time (s) under the standard all-query protocol. `Extra' denotes additional time over the raw backbone. For each method, we report the fastest stable implementation among the tested variants.}
    \begin{tabular}{l c c c c c c}
        \toprule[1pt]
        \multirow{2}{*}{\textbf{Method}} &
        \multicolumn{3}{c}{\textbf{Market1501}} &
        \multicolumn{3}{c}{\textbf{MSMT17}} \\
        \cmidrule(r){2-4}\cmidrule(l){5-7}
        & \textbf{Accel} & \textbf{Time} & \textbf{Extra}
        & \textbf{Accel} & \textbf{Time} & \textbf{Extra}\\
        \midrule
        Backbone & cuda & 24.5 & +0.0 & cuda & 145.7 & +0.0\\
        K-reciprocal~\cite{zhong2017re} & cuda-faiss & 31.6 & +7.1 & cuda-faiss & 184.4 & +38.7\\
        CA-Jaccard~\cite{chen2024jaccard} & cuda-faiss & 28.2 & +3.7 & cuda-faiss & 167.2 & +21.5\\
        ECN~\cite{sarfraz2018pose} & cuda & 25.0 & +0.5 & cuda-faiss & 147.5 & +1.7\\
        SFT~\cite{luo2019spectral} & cuda & 24.6 & +0.1 & cuda & 148.5 & +2.8\\
        AIM~\cite{fang2023visible} & cuda & 24.7 & +0.2 & cuda & 150.8 & +5.0\\
        NFormer~\cite{wang2022nformer} & cuda-sparse & 25.2 & +0.7 & cuda-sparse & 148.6 & +2.9\\
        GCR~\cite{zhang2023graph} & cuda-sparse & 25.2 & +0.7 & cuda-sparse & 159.3 & +13.6\\
        Cheb-GR~\cite{yang2025cheb} & cuda-dense & 25.3 & +0.8 & cuda-sparse & 361.8 & +216.1\\
        ANFI (Ours) & cuda-sparse & 24.8 & +0.3 & cuda-sparse & 148.5 & +2.7\\
        \bottomrule[1pt]
    \end{tabular}
    \vspace{-2mm}
    \label{tab:time_overhead}
\end{table}

To provide a controlled efficiency comparison, all gallery-based methods are evaluated with the same backbone features, runtime environment, and hardware under a unified PyTorch implementation with shared acceleration strategies, including GPU tensor computation, chunked evaluation to avoid dense full-matrix bottlenecks, and optional Faiss-based neighbor search.
All experiments in this section are conducted on a server with dual Intel(R) Xeon(R) Gold 5220R CPUs and NVIDIA GeForce RTX 3090 GPUs, and we report the fastest stable implementation for each method under this unified setting.

As shown in~\cref{tab:time_overhead}, the additional latency of gallery-based methods remains moderate compared with the shared raw-backbone extraction cost.
To reduce small run-to-run fluctuations in feature extraction, the reported total time uses a shared backbone extraction baseline plus the measured method-specific evaluation time.
On Market1501, all methods incur only limited extra cost over the backbone baseline.
On MSMT17, the relative ranking is similar, while the larger gallery size makes the latency differences more visible.
Among the compared methods, ANFI keeps competitive practical runtime while explicitly modeling discrepancy relations.
Cheb-GR shows a much larger overhead on MSMT17 because its row-wise adaptive threshold graph retains substantially more effective edges than kNN-style methods on the large-scale gallery, making graph propagation much denser in practice.

\subsection{Hyperparameter Analysis}

\begin{figure}[!t]
    \centering
    \includegraphics[width=0.70\linewidth]{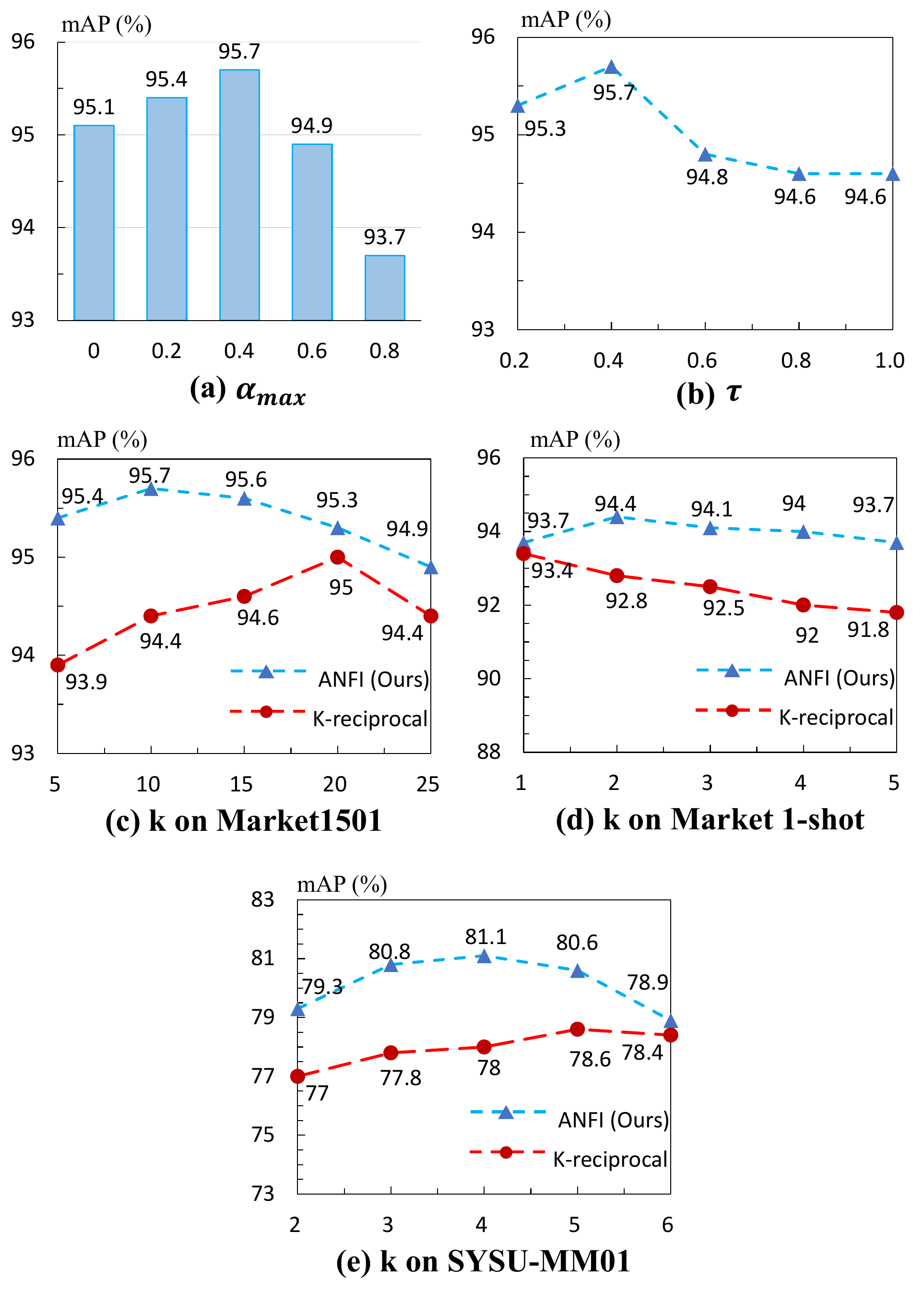}
    \caption{Effect of hyperparameters $\alpha_{max}$ and $\tau$ on Market1501, and $k$ on Market1501 and SYSU-MM01.}
    \hfill
    \vspace{-5mm}
    \label{fig:param}
\end{figure}

We analyzed the influences of key hyperparameters, including $\alpha_{max}$, $\tau$, and $k$.
As shown in~\cref{fig:param}(a), a smaller $\alpha_{max}$ cannot introduce sufficient noise, whereas excessive noise prevents the model from learning stable neighbor interactions.
The best performance is achieved at $\alpha_{max}=0.4$.
When using noise simulation with $\alpha_{max}=0.4$, we inject noisy neighbors during training to reduce the train-test gap caused by noisy neighbors.

As shown in~\cref{fig:param}(b), a smaller $\tau$ makes the model rely too heavily on self information, thereby underusing neighbors.
In contrast, a larger $\tau$ overemphasizes neighbors and increases sensitivity to false positive neighbors.
The setting $\tau=0.4$ provides the best trade-off.

As shown in~\cref{fig:param}(c), (d), and (e), the parameter $k$ is crucial for neighbor selection, and most neighbor-based or ranking-optimization methods involve a similar tuning parameter.
The optimal neighborhood size varies with the dataset and testing protocol.
In practice, we first determine a coarse candidate range of $k$ according to dataset protocol statistics, including the average number of gallery images per identity, and then tune $k$ for all compared methods within the same range under each setting.
To ensure fairness, we report each method with its best $k$ for each dataset and testing mode.
Furthermore, ANFI outperforms K-reciprocal~\cite{zhong2017re} across a wide range of $k$ values.

For discrepancy modeling, third-party samples $x_k$ are automatically selected as shared neighbors of two samples, i.e., $x_k\in\mathcal{N}_i\cap\mathcal{N}_j$.
Their number is determined by neighborhood overlap under $k$, without introducing additional hyperparameters.

\subsection{Visualizations}

\begin{figure}[!t]
    \centering
    \includegraphics[width=1.0\linewidth]{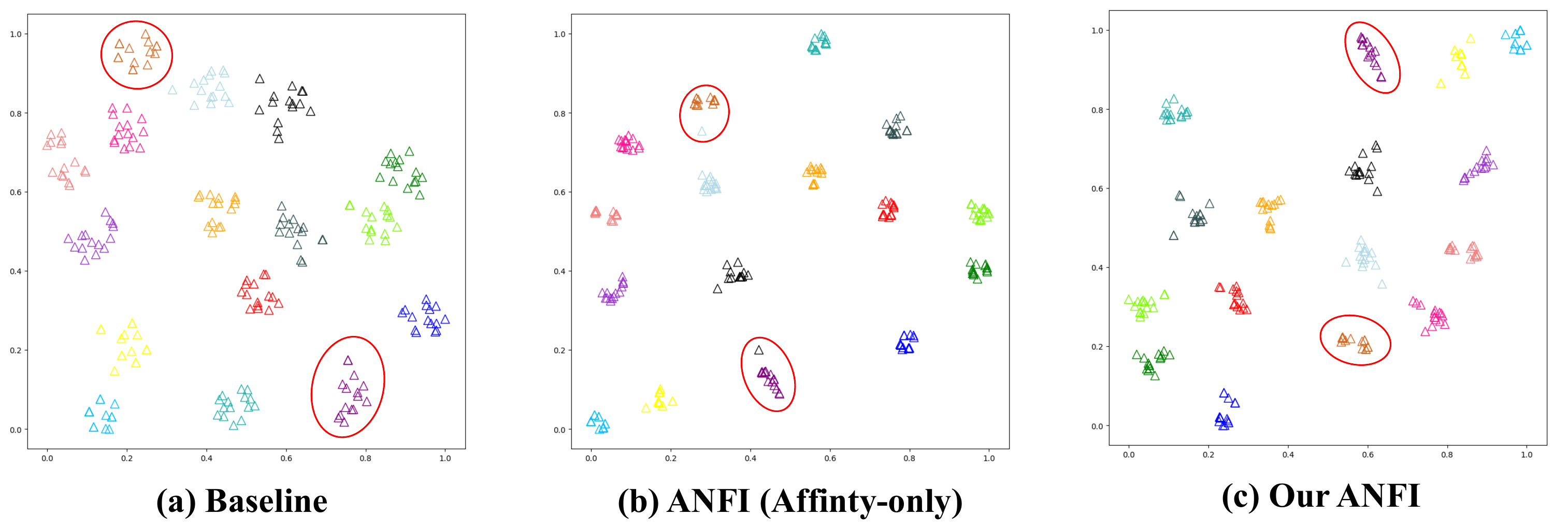}
    \caption{t-SNE visualization of the distributions of image features. Different colors represent different identities.}
    \label{fig:tsne}
    \vspace{-3mm}
\end{figure}

\cref{fig:tsne} presents t-SNE visualizations of feature distributions, where different colors represent different identities.
The baseline model exhibits substantial intra-class variance.
The affinity-only variant (ANFI without discrepancy relations) produces more compact intra-class clusters, but still tends to pull some false positive samples toward incorrect identities.
In contrast, ANFI further improves cluster compactness and separation, and reduces interference from false positive samples.

\begin{figure}[!t]
    \centering
    \includegraphics[width=0.8\linewidth]{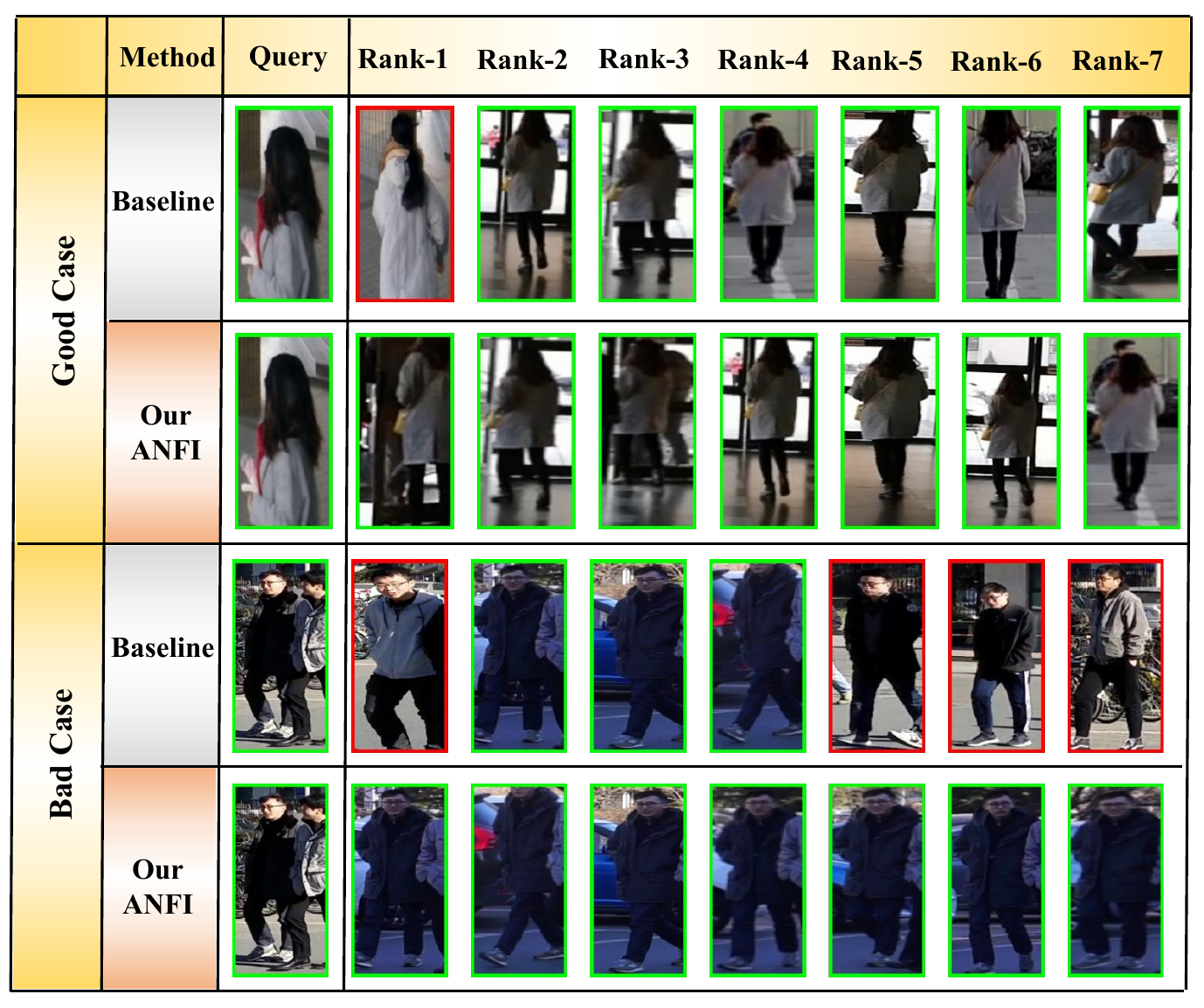}
    \caption{Visualization of retrieval ranking lists on the MSMT17 dataset. Green/red boxes represent correct/incorrect retrieval results.}
    \label{fig:rank}
    \vspace{-2mm}
\end{figure}

\cref{fig:rank} illustrates retrieval results on the large-scale MSMT17 dataset.
Green boxes indicate correct retrievals, while red boxes indicate incorrect retrievals.
The results in the 1$^{st}$ row correspond to a reliable neighbor distribution, where affinity learning dominates.
The 2$^{nd}$ row corresponds to a noisier neighbor distribution, where both affinity and discrepancy relations assist retrieval.
These results indicate that ANFI can handle different neighbor distributions and retrieve challenging positive targets that the baseline cannot find.
 
\section{Conclusion}

In this paper, we revisit neighbor-based Re-ID under varying neighbor reliability and show that affinity-only interaction is insufficient in the presence of noisy neighbors.
Based on this analysis, we propose Adaptive Neighbor Feature Interaction (ANFI), which jointly models affinity and discrepancy relations and adaptively mixes them for each sample.
We further introduce neighborhood similarity for discrepancy construction and Noisy Relation Supervision (NRS) to improve robustness under noisy neighbor distributions during training.
Extensive experiments across standard, cross-modal, and cross-domain settings, including comparisons with pure image-based, neighbor-based, and re-ranking methods, validate the effectiveness and robustness of ANFI.

\subsubsection*{Ethics Statement}
Person Re-ID can support public-safety and forensic search, but also has dual-use risks such as unauthorized tracking, mass surveillance, privacy invasion, and harms from misidentification. This work uses public benchmarks and does not collect new personal data or deploy a surveillance system. Real-world use should require legal authorization, data minimization, access control, auditing, and bias assessment.

\subsubsection*{Acknowledgements}
This work is supported by the National Natural Science Foundation of China (Grant No. 62272430).

\bibliographystyle{splncs04}
\bibliography{main}

@String(ECCV= {Eur. Conf. Comput. Vis.})

@String(AAAI = {AAAI})

@String(ECCV  = {ECCV})

@inproceedings{luo2019bag,
  title={Bag of tricks and a strong baseline for deep person re-identification},
  author={Luo, Hao and Gu, Youzhi and Liao, Xingyu and Lai, Shenqi and Jiang, Wei},
  booktitle={Proceedings of the IEEE/CVF conference on computer vision and pattern recognition workshops},
  pages={0--0},
  year={2019}
}

@article{ye2021deep,
  title={Deep learning for person re-identification: A survey and outlook},
  author={Ye, Mang and Shen, Jianbing and Lin, Gaojie and Xiang, Tao and Shao, Ling and Hoi, Steven CH},
  journal={IEEE transactions on pattern analysis and machine intelligence},
  volume={44},
  number={6},
  pages={2872--2893},
  year={2021},
  publisher={IEEE}
}

@inproceedings{wu2017rgb,
  title={RGB-infrared cross-modality person re-identification},
  author={Wu, Ancong and Zheng, Wei-Shi and Yu, Hong-Xing and Gong, Shaogang and Lai, Jianhuang},
  booktitle={Proceedings of the IEEE international conference on computer vision},
  pages={5380--5389},
  year={2017}
}

@inproceedings{zhong2020random,
  title={Random erasing data augmentation},
  author={Zhong, Zhun and Zheng, Liang and Kang, Guoliang and Li, Shaozi and Yang, Yi},
  booktitle={Proceedings of the AAAI conference on artificial intelligence},
  volume={34},
  number={07},
  pages={13001--13008},
  year={2020}
}

@inproceedings{li2022counterfactual,
  title={Counterfactual intervention feature transfer for visible-infrared person re-identification},
  author={Li, Xulin and Lu, Yan and Liu, Bin and Liu, Yating and Yin, Guojun and Chu, Qi and Huang, Jinyang and Zhu, Feng and Zhao, Rui and Yu, Nenghai},
  booktitle={European conference on computer vision},
  pages={381--398},
  year={2022},
  organization={Springer}
}

@inproceedings{lu2020cross,
  title={Cross-modality person re-identification with shared-specific feature transfer},
  author={Lu, Yan and Wu, Yue and Liu, Bin and Zhang, Tianzhu and Li, Baopu and Chu, Qi and Yu, Nenghai},
  booktitle={Proceedings of the IEEE/CVF conference on computer vision and pattern recognition},
  pages={13379--13389},
  year={2020}
}

@inproceedings{ye2020dynamic,
  title={Dynamic dual-attentive aggregation learning for visible-infrared person re-identification},
  author={Ye, Mang and Shen, Jianbing and J. Crandall, David and Shao, Ling and Luo, Jiebo},
  booktitle={European conference on computer vision},
  pages={229--247},
  year={2020},
  organization={Springer}
}

@inproceedings{luo2019spectral,
  title={Spectral feature transformation for person re-identification},
  author={Luo, Chuanchen and Chen, Yuntao and Wang, Naiyan and Zhang, Zhaoxiang},
  booktitle={Proceedings of the IEEE/CVF international conference on computer vision},
  pages={4976--4985},
  year={2019}
}

@inproceedings{zhong2017re,
  title={Re-ranking person re-identification with k-reciprocal encoding},
  author={Zhong, Zhun and Zheng, Liang and Cao, Donglin and Li, Shaozi},
  booktitle={Proceedings of the IEEE conference on computer vision and pattern recognition},
  pages={1318--1327},
  year={2017}
}

@article{zhang2020understanding,
  title={Understanding image retrieval re-ranking: A graph neural network perspective},
  author={Zhang, Xuanmeng and Jiang, Minyue and Zheng, Zhedong and Tan, Xiao and Ding, Errui and Yang, Yi},
  journal={arXiv preprint arXiv:2012.07620},
  year={2020}
}

@inproceedings{wang2022nformer,
  title={Nformer: Robust person re-identification with neighbor transformer},
  author={Wang, Haochen and Shen, Jiayi and Liu, Yongtuo and Gao, Yan and Gavves, Efstratios},
  booktitle={Proceedings of the IEEE/CVF conference on computer vision and pattern recognition},
  pages={7297--7307},
  year={2022}
}

@inproceedings{fang2023visible,
  title={Visible-infrared person re-identification via semantic alignment and affinity inference},
  author={Fang, Xingye and Yang, Yang and Fu, Ying},
  booktitle={Proceedings of the IEEE/CVF International Conference on Computer Vision},
  pages={11270--11279},
  year={2023}
}

@inproceedings{sun2018beyond,
  title={Beyond part models: Person retrieval with refined part pooling (and a strong convolutional baseline)},
  author={Sun, Yifan and Zheng, Liang and Yang, Yi and Tian, Qi and Wang, Shengjin},
  booktitle={Proceedings of the European conference on computer vision (ECCV)},
  pages={480--496},
  year={2018}
}

@inproceedings{wu2021discover,
  title={Discover cross-modality nuances for visible-infrared person re-identification},
  author={Wu, Qiong and Dai, Pingyang and Chen, Jie and Lin, Chia-Wen and Wu, Yongjian and Huang, Feiyue and Zhong, Bineng and Ji, Rongrong},
  booktitle={Proceedings of the IEEE/CVF conference on computer vision and pattern recognition},
  pages={4330--4339},
  year={2021}
}

@inproceedings{he2021transreid,
  title={Transreid: Transformer-based object re-identification},
  author={He, Shuting and Luo, Hao and Wang, Pichao and Wang, Fan and Li, Hao and Jiang, Wei},
  booktitle={Proceedings of the IEEE/CVF international conference on computer vision},
  pages={15013--15022},
  year={2021}
}

@inproceedings{sun2020circle,
  title={Circle loss: A unified perspective of pair similarity optimization},
  author={Sun, Yifan and Cheng, Changmao and Zhang, Yuhan and Zhang, Chi and Zheng, Liang and Wang, Zhongdao and Wei, Yichen},
  booktitle={Proceedings of the IEEE/CVF conference on computer vision and pattern recognition},
  pages={6398--6407},
  year={2020}
}

@inproceedings{kim2023partmix,
  title={Partmix: Regularization strategy to learn part discovery for visible-infrared person re-identification},
  author={Kim, Minsu and Kim, Seungryong and Park, Jungin and Park, Seongheon and Sohn, Kwanghoon},
  booktitle={Proceedings of the IEEE/CVF conference on computer vision and pattern recognition},
  pages={18621--18632},
  year={2023}
}

@inproceedings{ye2021channel,
  title={Channel augmented joint learning for visible-infrared recognition},
  author={Ye, Mang and Ruan, Weijian and Du, Bo and Shou, Mike Zheng},
  booktitle={Proceedings of the IEEE/CVF international conference on computer vision},
  pages={13567--13576},
  year={2021}
}

@inproceedings{zhang2023diverse,
  title={Diverse embedding expansion network and low-light cross-modality benchmark for visible-infrared person re-identification},
  author={Zhang, Yukang and Wang, Hanzi},
  booktitle={Proceedings of the IEEE/CVF conference on computer vision and pattern recognition},
  pages={2153--2162},
  year={2023}
}

@inproceedings{yang2023towards,
  title={Towards grand unified representation learning for unsupervised visible-infrared person re-identification},
  author={Yang, Bin and Chen, Jun and Ye, Mang},
  booktitle={Proceedings of the IEEE/CVF International Conference on Computer Vision},
  pages={11069--11079},
  year={2023}
}

@inproceedings{li2024adaptive,
  title={Adaptive high-frequency transformer for diverse wildlife re-identification},
  author={Li, Chenyue and Chen, Shuoyi and Ye, Mang},
  booktitle={European Conference on Computer Vision},
  pages={296--313},
  year={2024},
  organization={Springer}
}

@inproceedings{sarfraz2018pose,
  title={A pose-sensitive embedding for person re-identification with expanded cross neighborhood re-ranking},
  author={Sarfraz, M Saquib and Schumann, Arne and Eberle, Andreas and Stiefelhagen, Rainer},
  booktitle={Proceedings of the IEEE conference on computer vision and pattern recognition},
  pages={420--429},
  year={2018}
}

@inproceedings{chen2024jaccard,
  title={Ca-jaccard: camera-aware jaccard distance for person re-identification},
  author={Chen, Yiyu and Fan, Zheyi and Chen, Zhaoru and Zhu, Yixuan},
  booktitle={Proceedings of the IEEE/CVF conference on computer vision and pattern recognition},
  pages={17532--17541},
  year={2024}
}

@article{ye2016person,
  title={Person reidentification via ranking aggregation of similarity pulling and dissimilarity pushing},
  author={Ye, Mang and Liang, Chao and Yu, Yi and Wang, Zheng and Leng, Qingming and Xiao, Chunxia and Chen, Jun and Hu, Ruimin},
  journal={IEEE Transactions on Multimedia},
  volume={18},
  number={12},
  pages={2553--2566},
  year={2016},
  publisher={IEEE}
}

@inproceedings{zheng2015scalable,
  title={Scalable person re-identification: A benchmark},
  author={Zheng, Liang and Shen, Liyue and Tian, Lu and Wang, Shengjin and Wang, Jingdong and Tian, Qi},
  booktitle={Proceedings of the IEEE international conference on computer vision},
  pages={1116--1124},
  year={2015}
}

@inproceedings{wei2018person,
  title={Person transfer gan to bridge domain gap for person re-identification},
  author={Wei, Longhui and Zhang, Shiliang and Gao, Wen and Tian, Qi},
  booktitle={Proceedings of the IEEE conference on computer vision and pattern recognition},
  pages={79--88},
  year={2018}
}

@inproceedings{rao2021counterfactual,
  title={Counterfactual attention learning for fine-grained visual categorization and re-identification},
  author={Rao, Yongming and Chen, Guangyi and Lu, Jiwen and Zhou, Jie},
  booktitle={Proceedings of the IEEE/CVF international conference on computer vision},
  pages={1025--1034},
  year={2021}
}

@inproceedings{he2016deep,
  title={Deep residual learning for image recognition},
  author={He, Kaiming and Zhang, Xiangyu and Ren, Shaoqing and Sun, Jian},
  booktitle={Proceedings of the IEEE conference on computer vision and pattern recognition},
  pages={770--778},
  year={2016}
}

@inproceedings{li2014deepreid,
  title={Deepreid: Deep filter pairing neural network for person re-identification},
  author={Li, Wei and Zhao, Rui and Xiao, Tong and Wang, Xiaogang},
  booktitle={Proceedings of the IEEE conference on computer vision and pattern recognition},
  pages={152--159},
  year={2014}
}

@article{nguyen2017person,
  title={Person recognition system based on a combination of body images from visible light and thermal cameras},
  author={Nguyen, Dat Tien and Hong, Hyung Gil and Kim, Ki Wan and Park, Kang Ryoung},
  journal={Sensors},
  volume={17},
  number={3},
  pages={605},
  year={2017},
  publisher={MDPI}
}

@inproceedings{he2024instruct,
  title={Instruct-reid: A multi-purpose person re-identification task with instructions},
  author={He, Weizhen and Deng, Yiheng and Tang, Shixiang and Chen, Qihao and Xie, Qingsong and Wang, Yizhou and Bai, Lei and Zhu, Feng and Zhao, Rui and Ouyang, Wanli and others},
  booktitle={Proceedings of the IEEE/CVF Conference on Computer Vision and Pattern Recognition},
  pages={17521--17531},
  year={2024}
}

@inproceedings{li2023clip,
  title={Clip-reid: exploiting vision-language model for image re-identification without concrete text labels},
  author={Li, Siyuan and Sun, Li and Li, Qingli},
  booktitle={Proceedings of the AAAI conference on artificial intelligence},
  volume={37},
  number={1},
  pages={1405--1413},
  year={2023}
}

@article{zheng2024versatile,
  title={A versatile framework for multi-scene person re-identification},
  author={Zheng, Wei-Shi and Yan, Junkai and Peng, Yi-Xing},
  journal={IEEE Transactions on Pattern Analysis and Machine Intelligence},
  volume={47},
  number={3},
  pages={1362--1380},
  year={2024},
  publisher={IEEE}
}

@inproceedings{yang2025cheb,
  title={Cheb-GR: Rethinking k-nearest neighbor search in Re-ranking for Person Re-identification},
  author={Yang, Jinxi and Li, He and Du, Bo and Ye, Mang},
  booktitle={Proceedings of the IEEE/CVF Conference on Computer Vision and Pattern Recognition},
  pages={19261--19270},
  year={2025}
}

@article{zhang2023graph,
  title={Graph convolution based efficient re-ranking for visual retrieval},
  author={Zhang, Yuqi and Qian, Qi and Wang, Hongsong and Liu, Chong and Chen, Weihua and Wang, Fan},
  journal={IEEE Transactions on Multimedia},
  volume={26},
  pages={1089--1101},
  year={2023},
  publisher={IEEE}
}

@inproceedings{jiang2024domain,
  title={Domain shifting: A generalized solution for heterogeneous cross-modality person re-identification},
  author={Jiang, Yan and Cheng, Xu and Yu, Hao and Liu, Xingyu and Chen, Haoyu and Zhao, Guoying},
  booktitle={European Conference on Computer Vision},
  pages={289--306},
  year={2024},
  organization={Springer}
}

@inproceedings{li2025ATreid,
  title     = {Towards Anytime Retrieval: A Benchmark for Anytime Person Re-Identification},
  author    = {Li, Xulin and Lu, Yan and Liu, Bin and Li, Jiaze and Yang, Qinhong and Gong, Tao and Chu, Qi and Ye, Mang and Yu, Nenghai},
  booktitle = {Proceedings of the Thirty-Fourth International Joint Conference on Artificial Intelligence, {IJCAI-25}},
  publisher = {International Joint Conferences on Artificial Intelligence Organization},
  pages     = {1467--1475},
  year      = {2025},
}

@article{li2026causal,
  title={Causal Clothes-Invariant Feature Learning for Cloth-Changing Person Re-ID},
  author={Li, Xulin and Lu, Yan and Liu, Bin and Li, Jiaze and Liu, Yating and Chu, Qi and Ye, Mang and Ouyang, Wanli and Yu, Nenghai},
  journal={IEEE Transactions on Circuits and Systems for Video Technology},
  year={2026},
  publisher={IEEE}
}

@inproceedings{li2026mfen,
  title={MFEN: Multi-Frequency Expert Network for Visible-Infrared Person Re-ID},
  author={Li, Xulin and Lu, Yan and Liu, Bin and Yang, Qinhong and Chu, Qi and Gong, Tao and Yu, Nenghai},
  booktitle={Proceedings of the IEEE/CVF Conference on Computer Vision and Pattern Recognition},
  pages={18471--18480},
  year={2026}
}
\end{document}